\documentclass[11pt,letterpaper]{article}
\usepackage{ijcnlp2017}
\usepackage{times}
\usepackage{latexsym}
\usepackage{url}
\usepackage{fixltx2e}
\usepackage{amsmath}
\usepackage{blindtext}
\usepackage{enumitem}
\usepackage{dblfloatfix}
\usepackage{multirow}
\usepackage{graphicx}
\usepackage{endnotes}
\usepackage{dirtytalk}
\usepackage[export]{adjustbox}
\usepackage{caption, subcaption, floatrow}
\usepackage{setspace}

\ijcnlpfinalcopy

\def\ijcnlppaperid{207}

\newcommand{\mylinespacing}{0.920}

\title{Building a Neural Machine Translation System\\Using Only Synthetic Parallel Data}

\author{Jaehong Park \and Jongyoon Song \and Sungroh Yoon \\
Department of Electrical and Computer Engineering, Seoul National University \\
\texttt{\{uwanggood, coms1580, sryoon\}@snu.ac.kr}}

\date{}

\begin{document}

\maketitle

\begin{spacing}{\mylinespacing}
\begin{abstract}
Recent works have shown that synthetic parallel data automatically generated by translation models can be effective for various neural machine translation (NMT) issues. In this study, we build NMT systems using only synthetic parallel data. As an efficient alternative to real parallel data, we also present a new type of synthetic parallel corpus. The proposed pseudo parallel data are distinct from previous works in that ground truth and synthetic examples are mixed on both sides of sentence pairs. Experiments on Czech-German and French-German translations demonstrate the efficacy of the proposed pseudo parallel corpus, which shows not only enhanced results for bidirectional translation tasks but also substantial improvement with the aid of a ground truth real parallel corpus.
\end{abstract}

\section{Introduction}

Given the data-driven nature of neural machine translation (NMT), the limited source-to-target bilingual sentence pairs have been one of the major obstacles in building competitive NMT systems. Recently, pseudo parallel data, which refer to the synthetic bilingual sentence pairs automatically generated by existing translation models, have reported promising results with regard to the data scarcity in NMT. Many studies have found that the pseudo parallel data combined with the real bilingual parallel corpus significantly enhance the quality of NMT models~\cite{sennrich2015improving,zhang2016exploiting,cheng2016semi}. In addition, synthesized parallel data have played vital roles in many NMT problems such as domain adaptation~\cite{sennrich2015improving}, zero-resource NMT~\cite{firat2016zero}, and the rare word problem~\cite{zhang2016bridging}.

Inspired by their efficacy, we attempt to train NMT models using only synthetic parallel data. To the best of our knowledge, building NMT systems with only pseudo parallel data has yet to be studied. Through our research, we explore the availability of synthetic parallel data as an effective alternative to the real-world parallel corpus. The active usage of synthetic data in NMT particularly has its significance in low-resource environments where the ground truth parallel corpora are very limited or not established. Even in recent approaches such as zero-shot NMT~\cite{johnson2016google} and pivot-based NMT~\cite{cheng2016neural}, where direct source-to-target bilingual data are not required, the direct parallel corpus brings substantial improvements in translation quality where the pseudo parallel data can also be employed.

\begin{figure*}[t]
	\centering
	\includegraphics[width=\linewidth]{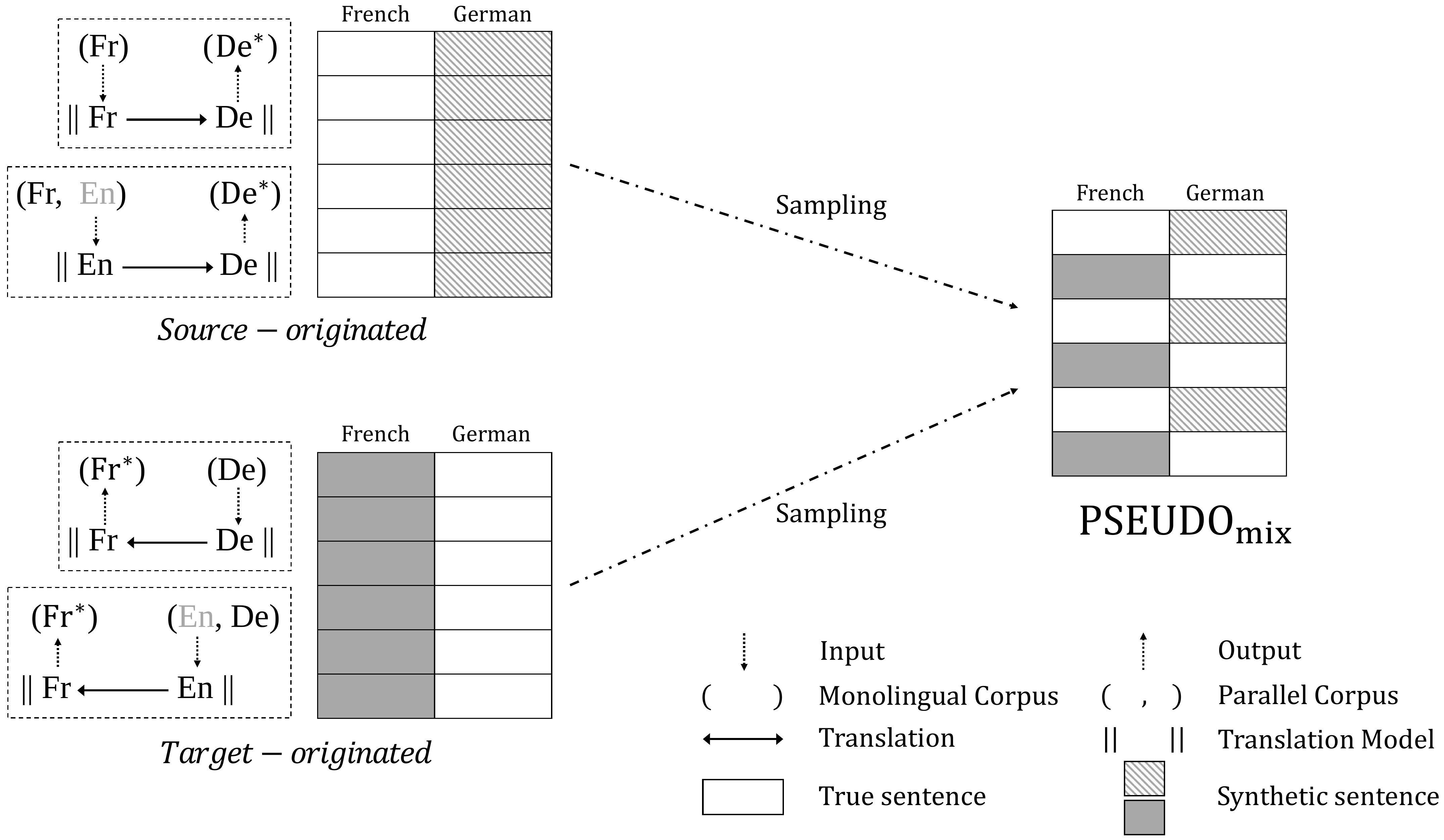}
    \caption{The process of building each pseudo parallel corpus group for French  \(\rightarrow\) German translation. * indicates the synthetic sentences generated by translation models. Each of the source-originated and the target-originated synthetic parallel data can be made from French or German monolingual corpora. They can also be built from parallel corpora including English, which is the pivot language.}
	\label{fig:pseudo_overview}
\end{figure*}

Previously suggested synthetic data, however, have several drawbacks to be a reliable alternative to the real parallel corpus. As illustrated in Figure~\ref{fig:pseudo_overview}, existing pseudo parallel corpora can be classified into two groups: \textit{source-originated} and \textit{target-originated}. The common property between them is that ground truth examples exist only on a single side (source or target) of pseudo sentence pairs, while the other side is composed of synthetic sentences only. The bias of synthetic examples in sentence pairs, however, may lead to the imbalance of the quality of learned NMT models when the given pseudo parallel corpus is exploited in bidirectional translation tasks (e.g., French\(\rightarrow\)German and German\(\rightarrow\)French). In addition, the reliability of the synthetic parallel data is heavily influenced by a single translation model where the synthetic examples originate. Low-quality synthetic sentences generated by the translation model would prevent NMT models from learning solid parameters.

To overcome these shortcomings, we propose a novel synthetic parallel corpus called PSEUDO\textsubscript{mix}. In contrast to previous works, PSEUDO\textsubscript{mix} includes both synthetic and real sentences on either side of sentence pairs. In practice, it can be readily built by mixing source- and target-originated pseudo parallel corpora for a given translation task. Experiments on several language pairs demonstrate that the proposed PSEUDO\textsubscript{mix} shows useful properties that make it a reliable candidate for real-world parallel data. In detail, we make the following contributions:

\begin{enumerate}

\item PSEUDO\textsubscript{mix} shows more balanced translation quality compared to existing pseudo parallel corpora in bidirectional translation tasks. For each task, it outperforms both source- and target-originated data when their performance gap is under a certain range.

\item When fine-tuned using real parallel data, the model trained with PSEUDO\textsubscript{mix} outperforms other fine-tuned models trained with source-originated and target-originated synthetic parallel data, indicating substantial improvement in translation quality.

\end{enumerate}

\section{Neural Machine Translation}

Given a source sentence \(x = (x_1, \ldots , x_m)\) and its corresponding target sentence \(y= (y_1, \ldots , y_n)\), the NMT aims to model the conditional probability \(p(y|x)\) with a single large neural network. To parameterize the conditional distribution, recent studies on NMT employ the encoder-decoder architecture~\cite{kalchbrenner2013recurrent,cho2014learning,sutskever2014sequence}. Thereafter, the attention mechanism~\cite{bahdanau2014neural,luong2015effective} has been introduced and successfully addressed the quality degradation of NMT when dealing with long input sentences~\cite{cho2014properties}.

In this study, we use the attentional NMT architecture proposed by Bahdanau et al. \shortcite{bahdanau2014neural}. In their work, the encoder, which is a bidirectional recurrent neural network, reads the source sentence and generates a sequence of source representations \(\bf{h} =(\bf{h_1}, \ldots , \bf{h_m}) \). The decoder, which is another recurrent neural network, produces the target sentence one symbol at a time. The log conditional probability thus can be decomposed as follows:

\begin{equation} 
\log p(y|x) = \sum_{t=1}^{n} \log p(y_t|y_{<t}, x)
\end{equation}
where \(y_{<t}\) = (\(y_1, \ldots , y_{t-1}\)). As described in Equation (2), the conditional distribution of \(p(y_t|y_{<t}, x)\) is modeled as a function of the previously predicted output \(y_{t-1}\), the hidden state of the decoder \(s_t\), and the context vector \(c_t\).

\begin{equation}
p(y_t|y_{<t}, x) \propto \exp \{g(y_{t-1}, s_t, c_t)\}
\end{equation}
The context vector \(c_t\) is used to determine the relevant part of the source sentence to predict \(y_t\). It is computed as the weighted sum of source representations \(\bf{h_1}, \ldots , \bf{h_m}\). Each weight \(\alpha_{ti}\) for \(\bf{h_i}\) implies the probability of the target symbol \(y_t\) being aligned to the source symbol \(x_i\):

\begin{equation}
c_t = \sum_{i=1}^{m} \alpha_{ti} \bf{h_i}
\end{equation}
Given a sentence-aligned parallel corpus of size \(N\), the entire parameter \(\theta\) of the NMT model is jointly trained to maximize the conditional probabilities of all sentence pairs \({ \{(x^n, y^n)\} }_{ n=1 }^{ N }\):

\begin{equation}
\theta^* = \underset{\theta}{\arg\!\max} \sum_{n=1}^{N} \log p(y^{n}|x^{n})
\end{equation}
where \(\theta^*\) is the optimal parameter.

\section{Related Work}
In statistical machine translation (SMT), synthetic bilingual data have been primarily proposed as a means to exploit monolingual corpora. By applying a self-training scheme, the pseudo parallel data were obtained by automatically translating the source-side monolingual corpora~\cite{ueffing2007transductive,wu2008domain}. In a similar but reverse way, the target-side monolingual corpora were also employed to build the synthetic parallel data~\cite{bertoldi2009domain,lambert2011investigations}. The primary goal of these works was to adapt trained SMT models to other domains using relatively abundant in-domain monolingual data.

Inspired by the successful application in SMT, there have been efforts to exploit synthetic parallel data in improving NMT systems. Source-side~\cite{zhang2016exploiting}, target-side~\cite{sennrich2015improving} and both sides~\cite{cheng2016semi} of the monolingual data have been used to build synthetic parallel corpora. In their work, the pseudo parallel data combined with a real training corpus significantly enhanced the translation quality of NMT. In Sennrich et al., \shortcite{sennrich2015improving}, domain adaptation of NMT was achieved by fine-tuning trained NMT models using a synthetic parallel corpus. Firat et al. \shortcite{firat2016zero} attempted to build NMT systems without any direct source-to-target parallel corpus. In their work, the pseudo parallel corpus was employed in fine-tuning the target-specific attention mechanism of trained multi-way multilingual NMT \cite{firat2016multi} models, which enabled zero-resource NMT between the source and target languages. Lastly, synthetic sentence pairs have been utilized to enrich the training examples having rare or unknown translation lexicons~\cite{zhang2016bridging}.

\section{Synthetic Parallel Data as an Alternative to Real Parallel Corpus}

\subsection{Motivation}
As described in the previous section, synthetic parallel data have been widely used to boost the performance of NMT. In this work, we further extend their application by training NMT with only synthetic data. In certain language pairs or domains where the source-to-target real parallel corpora are very rare or even unprepared, the model trained with synthetic parallel data can function as an effective baseline model. Once the additional ground truth parallel corpus is established, the trained model can be improved by retraining or fine-tuning using the real parallel data.

\subsection{Limits of the Previous Approaches}
For a given translation task, we classify the existing pseudo parallel data into the following groups:

\begin{enumerate}[label=(\alph*)]

\item \textit{Source-originated}: The source sentences are from a real corpus, and the associated target sentences are synthetic. The corpus can be formed by automatically translating a source-side monolingual corpus into the target language~\cite{zhang2016bridging,zhang2016exploiting}. It can also be built from source-pivot bilingual data by introducing a pivot language. In this case, a pivot-to-target translation model is employed to translate the pivot language corpus into the target language. The generated target sentences paired with the original source sentences form a pseudo parallel corpus.

\item \textit{Target-originated}: The target sentences are from a real corpus, and the associated source sentences are synthetic. The corpus can be formed by back-translating a target-side monolingual corpus into the source language \cite{sennrich2015improving}. Similar to the source-originated case, it can be built from a pivot-target bilingual corpus using a pivot-to-source translation model~\cite{firat2016zero}.

\end{enumerate}

The process of building each synthetic parallel corpus is illustrated in Figure~\ref{fig:pseudo_overview}. As shown in Figure~\ref{fig:pseudo_overview}, the previous studies on pseudo parallel data share a common property: synthetic and ground truth sentences are biased on a single side of sentence pairs. In such a case where the synthetic parallel data are the only or major resource used to train NMT, this may severely limit the availability of the given pseudo parallel corpus. For instance, as will be demonstrated in our experiments, synthetic data showing relatively high quality in one translation task (e.g., French\(\rightarrow\)German) can produce poor results in the translation task of the reverse direction (German\(\rightarrow\)French).

Another drawback of employing synthetic parallel data in training NMT is that the capacity of the synthetic parallel corpus is inherently influenced by the mother translation model from which the synthetic sentences originate. Depending on the quality of the mother model, ill-formed or inaccurate synthetic examples could be generated, which would negatively affect the reliability of the resultant synthetic parallel data. In the previous study, Zhang and Zong \shortcite{zhang2016exploiting} bypassed this issue by freezing the decoder parameters while training with the minibatches of pseudo bilingual pairs made from a source language monolingual corpus. This scheme, however, cannot be applied to our scenario as the decoder network will remain untrained during the entire training process.

\subsection{Proposed Mixing Approach}
To overcome the limitations of the previously suggested pseudo parallel data, we propose a new type of synthetic parallel corpus called PSEUDO\textsubscript{mix}. Our approach is quite straightforward: For a given translation task, we first build both source-originated and target-originated pseudo parallel data. PSEUDO\textsubscript{mix} can then be readily built by mixing them together. The overall process of building PSEUDO\textsubscript{mix} for the French\(\rightarrow\)German translation task is illustrated in Figure~\ref{fig:pseudo_overview}.

By mixing source- and target-originated pseudo parallel data, the resultant corpus includes both real and synthetic examples on either side of sentence pairs, which is the most evident feature of PSEUDO\textsubscript{mix}. Through the mixing approach, we attempt to lower the overall discrepancy in the quality of the source and target examples of synthetic sentence pairs, thus enhancing the reliability as a parallel resource. In the following section, we evaluate the actual benefits of the mixed composition in the synthetic parallel data.

\begin{table}[t]
\begin{center}
\begin{tabular}{c|c|cc}
		\hline 
			\multirow{2}*{Corpus} & \multirow{2}*{Size} & \multicolumn{2}{c}{Avg len} \\ 
				& & Fr & De \\ \cline{1-4}
                Europarl Fr-En-De & 1.78M & 26.00 & 23.16 \\
                Fr-De* & 1.45M & 25.56 & 22.98 \\
                Fr*-De & 1.45M & 25.32 & 23.46 \\
                PSEUDO\textsubscript{mix} & 1.45M & 25.47 & 23.26 \\
            \hline
		\end{tabular}
\end{center}
\caption{Statistics of the parallel corpora for Fr \(\leftrightarrow\) De translation tasks. The notation * denotes the synthetic part of the parallel corpus.}
\label{tab:training_corpora_frde_1}
\end{table}

\begin{table*}[t]
\begin{center}
\begin{tabular}{c||ccc||ccc}
		\hline 
        	\multirow{2}*{Corpus} & \multicolumn{3}{c||}{Fr \(\rightarrow\) De} & \multicolumn{3}{c}{De \(\rightarrow\) Fr} \\
			& \small{newstest2011} & \small{newstest2012} & \small{newstest2013} & \small{newstest2011} & \small{newstest2012} & \small{newstest2013} \\ \hline
            Fr-De* & 13.30 & 13.81 & 14.89 & 18.78 & 19.01 & 20.32 \\
            Fr*-De & 13.81 & \textbf{14.52} & 15.20 & 18.46 & 18.73 & 19.82 \\
            PSEUDO\textsubscript{mix} & \textbf{13.90} & 14.50 & \textbf{15.57} & \textbf{18.81} & \textbf{19.33} & \textbf{20.41} \\
            \hline
		\end{tabular}
\end{center}
\caption{Translation results (BLEU) for Fr \(\leftrightarrow\) De experiments. The notation * denotes the synthetic part of the parallel corpus. The highest BLEU for each set is bold-faced. }
\label{tab:pseudo_scratch_fr-de_1}
\end{table*}

\section{Experiments: Effects of Mixing Real and Synthetic Sentences}
In this section, we analyze the effects of the mixed composition in the synthetic parallel data. Mixing pseudo parallel corpora derived from different sources, however, inevitably brings diversity, which affects the capacity of the resulting corpus. We isolate this factor by building both source- and target-originated synthetic corpora from the identical source-to-target real parallel corpus. Our experiments are performed on French (Fr) \(\leftrightarrow\) German (De) translation tasks. Throughout the remaining paper, we use the notation * to denote the synthetic part of the pseudo sentence pairs.

\subsection{Data Preparation}
By choosing English (En) as the pivot language, we perform pivot alignments for identical English segments on Europarl Fr-En and En-De parallel corpora~\cite{koehn2005europarl}, constructing a multi-parallel corpus of Fr-En-De. Then each of the Fr*-De and Fr-De* pseudo parallel corpora is established from the multi-parallel data by applying the pivot language-based translation described in the previous section. For automatic translation, we utilize a pre-trained and publicly released NMT model~\footnote{ \url{http://data.statmt.org/rsennrich/wmt16_systems}} for En\(\rightarrow\)De and train another NMT model for En\(\rightarrow\)Fr using the WMT'15 En-Fr parallel corpus~\cite{bojar-EtAl:2015:WMT}. A beam of size 5 is used to generate synthetic sentences. Lastly, to match the size of the training data, PSEUDO\textsubscript{mix} is established by randomly sampling half of each Fr*-De and Fr-De* corpus and mixing them together.

\subsection{Data Preprocessing}
Each training corpus is tokenized using the tokenization script in Moses~\cite{koehn2007moses}. We represent every sentence as a sequence of subword units learned from byte-pair encoding~\cite{sennrich2015neural}. We remove empty lines and all the sentences of length over 50 subword units. For a fair comparison, all cleaned synthetic parallel data have equal sizes. The summary of the final parallel corpora is presented in Table~\ref{tab:training_corpora_frde_1}.

\subsection{Training and Evaluation}
All networks have 1024 hidden units and 500 dimensional embeddings. The vocabulary size is limited to 30K for each language. Each model is trained for 10 epochs using stochastic gradient descent with Adam~\cite{kingma2014adam}. The Minibatch size is  80, and the training set is reshuffled between every epoch. The norm of the gradient is clipped not to exceed 1.0~\cite{pascanu2013construct}. The learning rate is \(2 \cdot 10^{-4}\) in every case.

We use the newstest 2012 set for a development set and the newstest 2011 and newstest 2013 sets as test sets. At test time, beam search is used to approximately find the most likely translation. We use a beam of size 12 and normalize probabilities by the length of the candidate sentences. The evaluation metric is case-sensitive tokenized BLEU \cite{papineni2002bleu} computed with the \texttt{multi-bleu.perl} script from Moses. For each case, we present average BLEU evaluated on three different models trained from scratch.

\begin{table*}[t]
\begin{center}
\begin{tabular}{c||ccc||ccc}
	\hline 
       	\multirow{2}*{Corpus} & \multicolumn{3}{c||}{Fr \(\rightarrow\) De} & \multicolumn{3}{c}{De \(\rightarrow\) Fr} \\
		& \small{newstest2011} & \small{newstest2012} & \small{newstest2013} & \small{newstest2011} & \small{newstest2012} & \small{newstest2013} \\ \hline
           (a) Fr*-De \footnotesize{\textit{(K=3)}} & 13.76 & 14.43 & 15.18 & - & - & - \\ 
           (b) Fr*-De \footnotesize{\textit{(K=5)}} & 13.78 & \textbf{14.49} & 15.23 & 17.76 & 18.63 & 19.73 \\ 
           (a) + (b) & 13.74 & 14.38 & 15.27 & - & - & - \\ \hline
           (c) Fr-De* \footnotesize{\textit{(K=3)}} & - & - & - & 18.44 & 18.70 & 20.32 \\ 
           (d) Fr-De* \footnotesize{\textit{(K=5)}} & 13.36 & 14.08 & 15.28 & 18.18 & 18.76 & 20.13 \\
           (c) + (d) & - & - & - & 18.06 & 18.63 & 20.21 \\ \hline
           (b) + (d) & \textbf{13.93} & 14.27 & \textbf{15.53} & \textbf{18.52} & \textbf{19.04} & \textbf{20.33} \\ \hline
           
\end{tabular}
\end{center}
\caption{Translation results (BLEU) for Fr \(\leftrightarrow\) De experiments. \textit{K} denotes the beam size used to generate the corresponding synthetic parallel data. The highest BLEU for each set is bold-faced.}
\label{tab:reason_of_improvement}
\end{table*}

\subsection{Results and Analysis}

\subsubsection{A Comparison between Pivot-based Approach and Back-translation}
Before we choose the pivot language-based method for data synthesis, we conduct a preliminary experiment analyzing both pivot-based and direct back-translation. The model used for direct back-translation was trained with the ground truth Europarl Fr-De data made from the multi-parallel corpus presented in Table~\ref{tab:pseudo_scratch_fr-de_1}. On the newstest 2012/2013 sets, the synthetic corpus generated using the pivot approach showed higher BLEU (19.11 / 20.45) than the back-translation counterpart (18.23 / 19.81) when used in training a De\(\rightarrow\)Fr NMT model. Although the back-translation method has been effective in many studies~\cite{sennrich2015improving,sennrich2016edinburgh}, its availability becomes restricted in low-resource cases which is our major concern. This is due to the poor quality of the back-translation model built from the limited source-to-target parallel corpus. Instead, one can utilize abundant pivot-to-target parallel corpora by using a rich-resource language as the pivot language. This consequently improves the reliability of the quality of baseline translation models used for generating synthetic corpora.

\subsubsection{Effects of Mixing Source- and Target-originated Synthetic Data}
From Table~\ref{tab:pseudo_scratch_fr-de_1}, we find that the bias of the synthetic examples in pseudo parallel corpora brings imbalanced quality in the bidirectional translation tasks. Given that the source- and target-originated classification of a specific synthetic corpus is reversed depending on the direction of the translation, the overall results imply that the target-originated corpus for each translation task outperforms the source-originated data. The preference of target-originated synthetic data over the source-originated counterparts was formerly investigated in SMT by Lambert et al.,~\shortcite{lambert2011investigations}. In NMT, it can be explained by the degradation in quality in the source-originated data owing to the erroneous target language model formed by the synthetic target sentences. In contrast, we observe that PSEUDO\textsubscript{mix} not only produces balanced results for both Fr\(\rightarrow\)De and De\(\rightarrow\)Fr translation tasks but also shows the best or competitive translation quality for each task.

We note that mixing two different synthetic corpora leads to improved BLEU not their intermediate value. To investigate the cause of the improvement in PSEUDO\textsubscript{mix}, we build additional target-originated synthetic corpora for each Fr\(\leftrightarrow\)De translation with a beam of size 3. As shown in Table~\ref{tab:reason_of_improvement}, for the De\(\rightarrow\)Fr task, the new target-originated corpus (c) shows higher BLEU than the source-originated corpus (b) by itself. The improvement in BLEU, however, occurs only when mixing the source- and target-originated synthetic parallel data (b+d) compared to mixing two target-originated synthetic corpora (c+d). The same phenomenon is observed in the Fr\(\rightarrow\)De case as well. The results suggest that real and synthetic sentences mixed on either side of sentence pairs enhance the capability of a synthetic parallel corpus. We conjecture that ground truth examples in both encoder and decoder networks not only compensate for the erroneous language model learned from synthetic sentences but also reinforces patterns of use latent in the pseudo sentences.

\begin{table}[t]
\begin{center}
\begin{tabular}{c||cc|cc}
		\hline 
        	\multirow{2}*{Corpus} & \multicolumn{2}{c|}{Fr \(\rightarrow\) De} & \multicolumn{2}{c}{De \(\rightarrow\) Fr} \\
			& NMT & SMT & NMT & SMT \\ \cline{1-5}
            Fr-De* & 14.89 & 11.65 & 20.32 & 17.46 \\
            Fr*-De & 15.20 & 12.06 & 19.82 & 17.38 \\
            PSEUDO\textsubscript{mix} & \textbf{15.57} & \textbf{12.19} & \textbf{20.41} & \textbf{17.79} \\
            \hline
		\end{tabular}
\end{center}
\caption{Translation results (BLEU) for Fr \(\leftrightarrow\) De experiments evaluated on the newstest 2013 set.}
\label{tab:nmt_pbsmt}
\end{table}

\subsubsection{A Comparison with Phrase-based Statistical Machine Translation}
We also evaluate the effects of the proposed mixing strategy in phrase-based statistical machine translation~\cite{koehn2003statistical}. We use Moses~\cite{koehn2007moses} and its baseline configuration for training. A 5-gram Kneser-Ney model is used as the language model. Table~\ref{tab:nmt_pbsmt} shows the translation results of the phrase-based statistical machine translation (PBSMT) systems. In all experiments, NMT shows higher BLEU (2.44-3.38) compared to the PBSMT setting. We speculate that the deep architecture of NMT provides noise robustness in the synthetic examples. It is also notable that the proposed PSEUDO\textsubscript{mix} outperforms other synthetic corpora in PBSMT. The results clearly show that the benefit of the mixed composition in synthetic sentence pairs is beyond a specific machine translation framework.

\begin{table*}[t]
\centering
	\begin{subtable}[t]{.5\linewidth}
		\centering
		\begin{tabular}{c|c|cc}
			\hline 
                \multirow{2}*{Corpus} & \multirow{2}*{Size} & \multicolumn{2}{c}{Avg length} \\
                & & Cs & De \\ \cline{1-4}
                Europarl+NC11 & 0.6M & 23.54 & 25.49 \\ 
                Cs-De* & 3.5M & 25.33 & 26.01 \\
                Cs*-De & 3.5M & 23.31 & 25.37 \\
                PSEUDO\textsubscript{mix} & 3.5M & 24.39 & 25.72 \\
            \hline
		\end{tabular}
		\label{tab:training_corpora_csde}
        \caption{Cs \(\leftrightarrow\) De}
	\end{subtable}%
	\begin{subtable}[t]{.5\linewidth}
    	\centering
    	\begin{tabular}{c|c|cc}
        	\hline 
                \multirow{2}*{Corpus} & \multirow{2}*{Size} & \multicolumn{2}{c}{Avg length} \\
                & & Fr & De \\ \cline{1-4}
                Europarl+NC11 & 1.8M & 26.18 & 24.08 \\ 
                Fr-De* & 3.7M & 26.67 & 23.71 \\
                Fr*-De & 3.7M & 25.42 & 24.90 \\
                PSEUDO\textsubscript{mix} & 3.7M & 26.01 & 24.33 \\
            \hline
        \end{tabular}
		\label{tab:training_corpora_frde}
        \caption{Fr \(\leftrightarrow\) De}
    \end{subtable}	
\caption{Statistics of the training parallel corpora for large-scale Cs\(\leftrightarrow\)De and Fr\(\leftrightarrow\)De translation tasks.}
\label{tab:training_corpora}
\end{table*}

\section{Experiments: Large-scale Application}
The experiments shown in the previous section verify the potential of PSEUDO\textsubscript{mix} as an efficient alternative to the real parallel data. The condition in the previous case, however, is somewhat artificial, as we deliberately match the sources of all pseudo parallel corpora. In this section, we move on to more practical and large-scale applications of synthetic parallel data. Experiments are conducted on Czech (Cs) \(\leftrightarrow\) German (De) and French (Fr) \(\leftrightarrow\) German (De) translation tasks.

\subsection{Application Scenarios}
We analyze the efficacy of the proposed mixing approach in the following application scenarios:

\begin{enumerate}[label=(\roman*)]
\item \textit{Pseudo Only}: This setting trains NMT models using only synthetic parallel data without any ground truth parallel corpus.

\item \textit{Real Fine-tuning}: Once the training of an NMT model is completed in the Pseudo Only manner, the model is fine-tuned using only a ground truth parallel corpus.

\end{enumerate}

The suggested scenarios reflect low-resource situations in building NMT systems. In the Real Fine-tuning, we fine-tune the best model of the Pseudo Only scenario evaluated on the development set.

\subsection{Data Preparation}
We use the parallel corpora from the shared translation task of WMT'15 and WMT'16~\cite{bojar-EtAl:2016:WMT1}. Using the same pivot-based technique as the previous task, Cs-De* and Fr-De* corpora are built from the WMT'15 Cs-En and Fr-En parallel data respectively. For Cs*-De and Fr*-De, WMT'16 En-De parallel data are employed. We again use pre-trained NMT models for En\(\rightarrow\)Cs, En\(\rightarrow\)De, and En\(\rightarrow\)Fr to generate synthetic sentences. A beam of size 1 is used for fast decoding.

For the Real Fine-tuning scenario, we use real parallel corpora from the Europarl and News Commentary11 dataset. These direct parallel corpora are obtained from OPUS~\cite{tiedemann2012parallel}. The size of each set of ground truth and synthetic parallel data is presented in Table~\ref{tab:training_corpora}. Given that the training corpus for widely studied language pairs amounts to several million lines, the Cs-De language pair (0.6M) reasonably represents a low-resource situation. On the other hand, the Fr-De language pair (1.8M) is considered to be relatively resource-rich in our experiments. The details of the preprocessing are identical to those in the previous case.

\subsection{Training and Evaluation}
We use the same experimental settings that we used for the previous case except for the Real Fine-tuning scenario. In the fine-tuning step, we use the learning rate of \(2 \cdot 10^{-5}\), which produced better results. Embeddings are fixed throughout the fine-tuning steps. For evaluation, we use the same development and test sets used in the previous task.

\subsection{Results and Analysis}

\subsubsection{A Comparison with Real Parallel Data}
Table~\ref{tab:real_finetuning} shows the results of the Pseudo Only scenario on Cs\(\leftrightarrow\)De and Fr\(\leftrightarrow\)De tasks. For the baseline comparison, we also present the translation quality of the NMT models trained with the ground truth Europarl+NC11 parallel corpora (a). In Cs\(\leftrightarrow\)De, the Pseudo Only scenario shows outperforming results compared to the real parallel corpus by up to 3.86-4.43 BLEU on the newstest 2013 set. Even for the Fr\(\leftrightarrow\)De case, where the size of the real parallel corpus is relatively large, the best BLEU of the pseudo parallel corpora is higher than that of the real parallel corpus by 1.3 (Fr\(\rightarrow\)De) and 0.49 (De\(\rightarrow\)Fr). We list the results on the newstest 2011 and newstest 2012 in the appendix. From the results, we conclude that large-scale synthetic parallel data can perform as an effective alternative to the real parallel corpora, particularly in low-resource language pairs.

\begin{table*}[t]
\centering
	\begin{subtable}[t]{\columnwidth}
		\centering
		\begin{tabular}{c||cc||cc}
        		\hline
                \textbf{Baseline} & \multicolumn{2}{c||}{Cs \(\rightarrow\) De} & \multicolumn{2}{c}{De \(\rightarrow\) Cs} \\
				(a) Europarl+NC11 & \multicolumn{2}{c||}{14.96} & \multicolumn{2}{c}{12.36} \\ 
                (b) +Pivot back-trans corpus & \multicolumn{2}{c||}{{\scriptsize(+4.02)} 18.98} & \multicolumn{2}{c}{{\scriptsize(+4.40)} 16.76} \\ \hline\hline
				\textbf{Synthetic Corpus} & Pseudo Only & Real Fine-tuning & Pseudo Only & Real Fine-tuning \\
                Cs-De* & 16.87 & {\scriptsize(+1.95)} 18.82 & 15.29 & {\scriptsize(+1.21)} 16.50 \\
                Cs*-De & 18.62 & {\scriptsize(+0.40)} 19.02 & 16.51 & {\scriptsize(+0.45)} 16.96 \\
                PSEUDO\textsubscript{mix} & 18.82 & {\scriptsize(+0.53)} \textbf{19.35} & 16.79 & {\scriptsize(+0.68)} \textbf{17.47} \\
            \hline
		\end{tabular}
		\label{tab:real_finetuning_csde}
        \caption{Cs \(\leftrightarrow\) De}
	\end{subtable}%
	\qquad
    \begin{subtable}[t]{\columnwidth}
		\centering
		\begin{tabular}{c||cc||cc}
        		\hline
                \textbf{Baseline} & \multicolumn{2}{c||}{Fr \(\rightarrow\) De} & \multicolumn{2}{c}{De \(\rightarrow\) Fr} \\
				(a) Europarl+NC11 & \multicolumn{2}{c||}{17.68} & \multicolumn{2}{c}{22.39} \\ 
                (b) +Pivot back-trans corpus & \multicolumn{2}{c||}{{\scriptsize(+1.59)} 19.27} & \multicolumn{2}{c}{{\scriptsize(+1.93)} 24.32} \\ \hline\hline
				\textbf{Synthetic Corpus} & Pseudo Only & Real Fine-tuning & Pseudo Only & Real Fine-tuning \\
                Fr-De* & 17.57 & {\scriptsize(+1.65)} 19.22 & 22.88 & {\scriptsize(+1.42)} 24.30 \\
                Fr*-De & 18.55 & {\scriptsize(+1.04)} 19.59 & 19.87 & {\scriptsize(+4.74)} 24.61 \\
                PSEUDO\textsubscript{mix} & 18.98 & {\scriptsize(+0.87)} \textbf{19.85} & 22.71 & {\scriptsize(+1.99)} \textbf{24.70} \\
            \hline
		\end{tabular}
		\label{tab:real_finetuning_frde}
        \caption{Fr \(\leftrightarrow\) De}
	\end{subtable}
    
\caption{Translation results (BLEU) for Pseudo Only and Real Fine-tuning scenarios evaluated on the newstest 2013 set. For the results of the Real Fine-tuning, the values in parentheses are improvements in BLEU compared to the Pseudo Only setting. The highest BLEU for each translation task is bold-faced.}
\label{tab:real_finetuning}
\end{table*}

\begin{figure}[t]
	\centering
    \includegraphics[width=\linewidth]{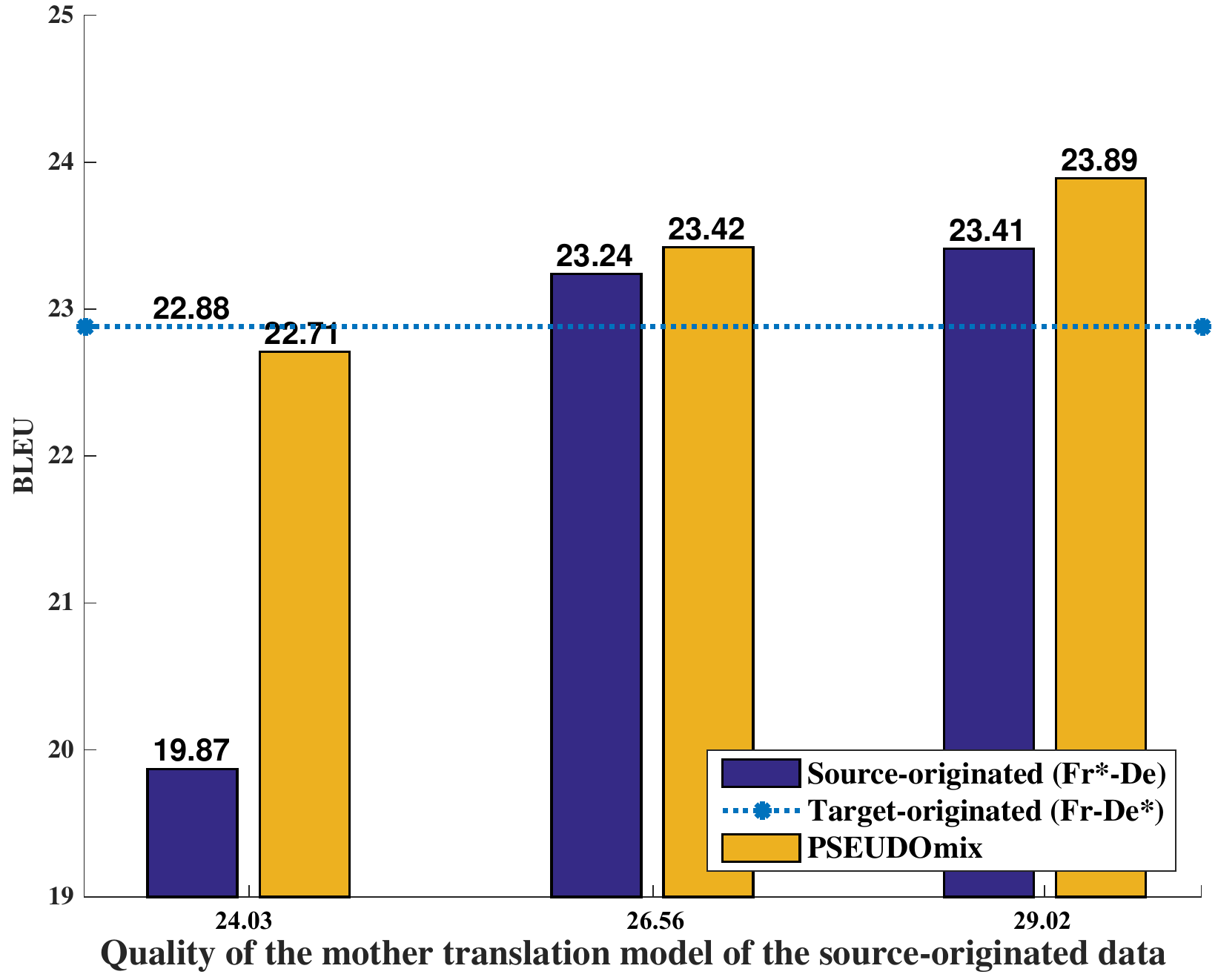}
    \caption{Translation results for the De \(\rightarrow\) Fr task on the newstest 2013 set with respect to the quality of the mother model for the source-originated Fr*-De data. The quality of the mother model is evaluated on the En-Fr newstest 2012 set.}
    \label{fig:effect_of_mother_model}
\end{figure}

\subsubsection{Results from the Pseudo Only Scenario}
As shown in Table~\ref{tab:real_finetuning}, the model learned from the Cs*-De corpus outperforms the model trained with the Cs-De* corpus in every case. This result is slightly different from the previous case, where the target-originated synthetic corpus for each translation task reports better results than the source-originated data. This arises from the diversity in the source of each pseudo parallel corpus, which vary in their suitability for the given test set. Table~\ref{tab:real_finetuning} also shows that mixing the Cs*-De corpus with the Cs-De* corpus of worse quality brings improvements in the resulting PSEUDO\textsubscript{mix}, showing the highest BLEU for bidirectional Cs\(\leftrightarrow\)De translation tasks. In addition, PSEUDO\textsubscript{mix} again shows much more balanced performance in Fr\(\leftrightarrow\)De translations compared to other synthetic parallel corpora.

While the mixing strategy compensates for most of the gap between the Fr-De* and the Fr*-De (3.01\(\rightarrow\)0.17) in the De\(\rightarrow\)Fr case, the resulting PSEUDO\textsubscript{mix} still shows lower BLEU than the target-originated Fr-De* corpus. We thus enhance the quality of the synthetic examples of the source-originated Fr*-De data by further training its mother translation model (En\(\rightarrow\)Fr). As illustrated in Figure~\ref{fig:effect_of_mother_model}, with the target-originated Fr-De* corpus being fixed, the quality of the models trained with the source-originated Fr*-De data and PSEUDO\textsubscript{mix} increases in proportion to the quality of the mother model for the Fr*-De corpus. Eventually, PSEUDO\textsubscript{mix} shows the highest BLEU, outperforming both Fr*-De and Fr-De* data. The results indicate that the benefit of the proposed mixing approach becomes much more evident when the quality gap between the source- and target-originated synthetic data is within a certain range.

\subsubsection{Results from the Real Fine-tuning Scenario}
As presented in Table~\ref{tab:real_finetuning}, we observe that fine-tuning using ground truth parallel data brings substantial improvements in the translation qualities of all NMT models. Among all fine-tuned models, PSEUDO\textsubscript{mix} shows the best performance in all experiments. This is particularly encouraging for the case of De\(\rightarrow\)Fr, where PSEUDO\textsubscript{mix} reported lower BLEU than the Fr-De* data before it was fine-tuned. Even in the case where PSEUDO\textsubscript{mix} shows comparable results with other synthetic corpora in the Pseudo Only scenario, it shows higher improvements in the translation quality when fine-tuned with the real parallel data. These results clearly demonstrate the strengths of the proposed PSEUDO\textsubscript{mix}, which indicate both competitive translation quality by itself and relatively higher potential improvement as a result of the refinement using ground truth parallel corpora.

In Table~\ref{tab:real_finetuning} (b), we also present the performance of NMT models learned from the ground truth Europarl+NC11 data merged with the target-originated synthetic parallel corpus for each task. This is identical in spirit to the method in Sennrich et al.~\shortcite{sennrich2015improving} which employs back-translation for data synthesis. Instead of direct back-translation, we used pivot-based back-translation, as we verified the strength of the pivot-based data synthesis in low-resource environments. Although the ground truth data is only used for the refinement, the Real Fine-tuning scheme applied to PSEUDO\textsubscript{mix} shows better translation quality compared to the models trained with the merged corpus (b). Even the results of the Real Fine-tuning on the target-originated corpus provide comparable results to the training with the merged corpus from scratch. The overall results support the efficacy of the proposed two-step methods in practical application: the Pseudo Only method to introduce useful prior on the NMT parameters and the Real Fine-tuning scheme to reorganize the pre-trained NMT parameters using in-domain parallel data.

\section{Conclusion}
In this work, we have constructed NMT systems using only synthetic parallel data. For this purpose, we suggest a novel pseudo parallel corpus called PSEUDO\textsubscript{mix} where synthetic and ground truth real examples are mixed on either side of sentence pairs. Experiments show that the proposed PSEUDO\textsubscript{mix} not only shows enhanced results for bidirectional translation but also reports substantial improvement when fine-tuned with ground truth parallel data. Our work has significance in that it provides a thorough investigation on the use of synthetic parallel corpora in low-resource NMT environment. Without any adjustment, the proposed method can also be extended to other learning areas where parallel samples are employed. For future work, we plan to explore robust data sampling methods, which would maximize the quality of the mixed synthetic parallel data.

\end{spacing}

\bibliography{ijcnlp2017}

\begin{thebibliography}{}
\expandafter\ifx\csname natexlab\endcsname\relax\def\natexlab#1{#1}\fi

\bibitem[{Bahdanau et~al.(2014)Bahdanau, Cho, and Bengio}]{bahdanau2014neural}
Dzmitry Bahdanau, Kyunghyun Cho, and Yoshua Bengio. 2014.
\newblock Neural machine translation by jointly learning to align and
  translate.
\newblock {\em arXiv preprint arXiv:1409.0473\/} .

\bibitem[{Bertoldi and Federico(2009)}]{bertoldi2009domain}
Nicola Bertoldi and Marcello Federico. 2009.
\newblock Domain adaptation for statistical machine translation with
  monolingual resources.
\newblock In {\em Proceedings of the fourth workshop on statistical machine
  translation\/}. Association for Computational Linguistics, pages 182--189.

\bibitem[{Bojar et~al.(2016)Bojar, Chatterjee, Federmann, Graham, Haddow, Huck,
  Jimeno~Yepes, Koehn, Logacheva, Monz, Negri, Neveol, Neves, Popel, Post,
  Rubino, Scarton, Specia, Turchi, Verspoor, and
  Zampieri}]{bojar-EtAl:2016:WMT1}
Ond\v{r}ej Bojar, Rajen Chatterjee, Christian Federmann, Yvette Graham, Barry
  Haddow, Matthias Huck, Antonio Jimeno~Yepes, Philipp Koehn, Varvara
  Logacheva, Christof Monz, Matteo Negri, Aurelie Neveol, Mariana Neves, Martin
  Popel, Matt Post, Raphael Rubino, Carolina Scarton, Lucia Specia, Marco
  Turchi, Karin Verspoor, and Marcos Zampieri. 2016.
\newblock \href{http://www.aclweb.org/anthology/W/W16/W16-2301}{Findings of the
  2016 conference on machine translation}.
\newblock In {\em Proceedings of the First Conference on Machine
  Translation\/}. Association for Computational Linguistics, Berlin, Germany,
  pages 131--198.
\newblock
  \href{http://www.aclweb.org/anthology/W/W16/W16-2301}{http://www.aclweb.org/anthology/W/W16/W16-2301}.

\bibitem[{Bojar et~al.(2015)Bojar, Chatterjee, Federmann, Haddow, Huck, Hokamp,
  Koehn, Logacheva, Monz, Negri, Post, Scarton, Specia, and
  Turchi}]{bojar-EtAl:2015:WMT}
Ond\v{r}ej Bojar, Rajen Chatterjee, Christian Federmann, Barry Haddow, Matthias
  Huck, Chris Hokamp, Philipp Koehn, Varvara Logacheva, Christof Monz, Matteo
  Negri, Matt Post, Carolina Scarton, Lucia Specia, and Marco Turchi. 2015.
\newblock \href{http://aclweb.org/anthology/W15-3001}{Findings of the 2015
  workshop on statistical machine translation}.
\newblock In {\em Proceedings of the Tenth Workshop on Statistical Machine
  Translation\/}. Association for Computational Linguistics, Lisbon, Portugal,
  pages 1--46.
\newblock
  \href{http://aclweb.org/anthology/W15-3001}{http://aclweb.org/anthology/W15-3001}.

\bibitem[{Cheng et~al.(2016{\natexlab{a}})Cheng, Liu, Yang, Sun, and
  Xu}]{cheng2016neural}
Yong Cheng, Yang Liu, Qian Yang, Maosong Sun, and Wei Xu. 2016{\natexlab{a}}.
\newblock Neural machine translation with pivot languages.
\newblock {\em arXiv preprint arXiv:1611.04928\/} .

\bibitem[{Cheng et~al.(2016{\natexlab{b}})Cheng, Xu, He, He, Wu, Sun, and
  Liu}]{cheng2016semi}
Yong Cheng, Wei Xu, Zhongjun He, Wei He, Hua Wu, Maosong Sun, and Yang Liu.
  2016{\natexlab{b}}.
\newblock Semi-supervised learning for neural machine translation.
\newblock {\em arXiv preprint arXiv:1606.04596\/} .

\bibitem[{Cho et~al.(2014{\natexlab{a}})Cho, van Merrienboer, Bahdanau, and
  Bengio}]{cho2014properties}
Kyunghyun Cho, Bart van Merrienboer, Dzmitry Bahdanau, and Yoshua Bengio.
  2014{\natexlab{a}}.
\newblock On the properties of neural machine translation: Encoder-decoder
  approaches.
\newblock In {\em Eighth Workshop on Syntax, Semantics and Structure in
  Statistical Translation (SSST-8)\/}.

\bibitem[{Cho et~al.(2014{\natexlab{b}})Cho, Van~Merri{\"e}nboer, Gulcehre,
  Bahdanau, Bougares, Schwenk, and Bengio}]{cho2014learning}
Kyunghyun Cho, Bart Van~Merri{\"e}nboer, Caglar Gulcehre, Dzmitry Bahdanau,
  Fethi Bougares, Holger Schwenk, and Yoshua Bengio. 2014{\natexlab{b}}.
\newblock Learning phrase representations using rnn encoder-decoder for
  statistical machine translation.
\newblock {\em arXiv preprint arXiv:1406.1078\/} .

\bibitem[{Firat et~al.(2016{\natexlab{a}})Firat, Cho, and
  Bengio}]{firat2016multi}
Orhan Firat, Kyunghyun Cho, and Yoshua Bengio. 2016{\natexlab{a}}.
\newblock Multi-way, multilingual neural machine translation with a shared
  attention mechanism.
\newblock {\em arXiv preprint arXiv:1601.01073\/} .

\bibitem[{Firat et~al.(2016{\natexlab{b}})Firat, Sankaran, Al-Onaizan, Vural,
  and Cho}]{firat2016zero}
Orhan Firat, Baskaran Sankaran, Yaser Al-Onaizan, Fatos T~Yarman Vural, and
  Kyunghyun Cho. 2016{\natexlab{b}}.
\newblock Zero-resource translation with multi-lingual neural machine
  translation.
\newblock {\em arXiv preprint arXiv:1606.04164\/} .

\bibitem[{Johnson et~al.(2016)Johnson, Schuster, Le, Krikun, Wu, Chen, Thorat,
  Vi{\'e}gas, Wattenberg, Corrado et~al.}]{johnson2016google}
Melvin Johnson, Mike Schuster, Quoc~V Le, Maxim Krikun, Yonghui Wu, Zhifeng
  Chen, Nikhil Thorat, Fernanda Vi{\'e}gas, Martin Wattenberg, Greg Corrado,
  et~al. 2016.
\newblock Google's multilingual neural machine translation system: Enabling
  zero-shot translation.
\newblock {\em arXiv preprint arXiv:1611.04558\/} .

\bibitem[{Kalchbrenner and Blunsom(2013)}]{kalchbrenner2013recurrent}
Nal Kalchbrenner and Phil Blunsom. 2013.
\newblock Recurrent continuous translation models.
\newblock In {\em EMNLP\/}. volume~3, page 413.

\bibitem[{Kingma and Ba(2014)}]{kingma2014adam}
Diederik Kingma and Jimmy Ba. 2014.
\newblock Adam: A method for stochastic optimization.
\newblock {\em arXiv preprint arXiv:1412.6980\/} .

\bibitem[{Koehn(2005)}]{koehn2005europarl}
Philipp Koehn. 2005.
\newblock Europarl: A parallel corpus for statistical machine translation.
\newblock In {\em MT summit\/}. volume~5, pages 79--86.

\bibitem[{Koehn et~al.(2007)Koehn, Hoang, Birch, Callison-Burch, Federico,
  Bertoldi, Cowan, Shen, Moran, Zens et~al.}]{koehn2007moses}
Philipp Koehn, Hieu Hoang, Alexandra Birch, Chris Callison-Burch, Marcello
  Federico, Nicola Bertoldi, Brooke Cowan, Wade Shen, Christine Moran, Richard
  Zens, et~al. 2007.
\newblock Moses: Open source toolkit for statistical machine translation.
\newblock In {\em Proceedings of the 45th annual meeting of the ACL on
  interactive poster and demonstration sessions\/}. Association for
  Computational Linguistics, pages 177--180.

\bibitem[{Koehn et~al.(2003)Koehn, Och, and Marcu}]{koehn2003statistical}
Philipp Koehn, Franz~Josef Och, and Daniel Marcu. 2003.
\newblock Statistical phrase-based translation.
\newblock In {\em Proceedings of the 2003 Conference of the North American
  Chapter of the Association for Computational Linguistics on Human Language
  Technology-Volume 1\/}. Association for Computational Linguistics, pages
  48--54.

\bibitem[{Lambert et~al.(2011)Lambert, Schwenk, Servan, and
  Abdul-Rauf}]{lambert2011investigations}
Patrik Lambert, Holger Schwenk, Christophe Servan, and Sadaf Abdul-Rauf. 2011.
\newblock Investigations on translation model adaptation using monolingual
  data.
\newblock In {\em Proceedings of the Sixth Workshop on Statistical Machine
  Translation\/}. Association for Computational Linguistics, pages 284--293.

\bibitem[{Luong et~al.(2015)Luong, Pham, and Manning}]{luong2015effective}
Minh-Thang Luong, Hieu Pham, and Christopher~D Manning. 2015.
\newblock Effective approaches to attention-based neural machine translation.
\newblock {\em arXiv preprint arXiv:1508.04025\/} .

\bibitem[{Papineni et~al.(2002)Papineni, Roukos, Ward, and
  Zhu}]{papineni2002bleu}
Kishore Papineni, Salim Roukos, Todd Ward, and Wei-Jing Zhu. 2002.
\newblock Bleu: a method for automatic evaluation of machine translation.
\newblock In {\em Proceedings of the 40th annual meeting on association for
  computational linguistics\/}. Association for Computational Linguistics,
  pages 311--318.

\bibitem[{Pascanu et~al.(2013)Pascanu, Gulcehre, Cho, and
  Bengio}]{pascanu2013construct}
Razvan Pascanu, Caglar Gulcehre, Kyunghyun Cho, and Yoshua Bengio. 2013.
\newblock How to construct deep recurrent neural networks.
\newblock {\em arXiv preprint arXiv:1312.6026\/} .

\bibitem[{Sennrich et~al.(2015{\natexlab{a}})Sennrich, Haddow, and
  Birch}]{sennrich2015improving}
Rico Sennrich, Barry Haddow, and Alexandra Birch. 2015{\natexlab{a}}.
\newblock Improving neural machine translation models with monolingual data.
\newblock {\em arXiv preprint arXiv:1511.06709\/} .

\bibitem[{Sennrich et~al.(2015{\natexlab{b}})Sennrich, Haddow, and
  Birch}]{sennrich2015neural}
Rico Sennrich, Barry Haddow, and Alexandra Birch. 2015{\natexlab{b}}.
\newblock Neural machine translation of rare words with subword units.
\newblock {\em arXiv preprint arXiv:1508.07909\/} .

\bibitem[{Sennrich et~al.(2016)Sennrich, Haddow, and
  Birch}]{sennrich2016edinburgh}
Rico Sennrich, Barry Haddow, and Alexandra Birch. 2016.
\newblock Edinburgh neural machine translation systems for wmt 16.
\newblock {\em arXiv preprint arXiv:1606.02891\/} .

\bibitem[{Sutskever et~al.(2014)Sutskever, Vinyals, and
  Le}]{sutskever2014sequence}
Ilya Sutskever, Oriol Vinyals, and Quoc~V Le. 2014.
\newblock Sequence to sequence learning with neural networks.
\newblock In {\em Advances in neural information processing systems\/}. pages
  3104--3112.

\bibitem[{Tiedemann(2012)}]{tiedemann2012parallel}
J{\"o}rg Tiedemann. 2012.
\newblock Parallel data, tools and interfaces in opus.
\newblock In {\em LREC\/}. volume 2012, pages 2214--2218.

\bibitem[{Ueffing et~al.(2007)Ueffing, Haffari, Sarkar
  et~al.}]{ueffing2007transductive}
Nicola Ueffing, Gholamreza Haffari, Anoop Sarkar, et~al. 2007.
\newblock Transductive learning for statistical machine translation.
\newblock In {\em Annual Meeting-Association for Computational Linguistics\/}.
  volume~45, page~25.

\bibitem[{Wu et~al.(2008)Wu, Wang, and Zong}]{wu2008domain}
Hua Wu, Haifeng Wang, and Chengqing Zong. 2008.
\newblock Domain adaptation for statistical machine translation with domain
  dictionary and monolingual corpora.
\newblock In {\em Proceedings of the 22nd International Conference on
  Computational Linguistics-Volume 1\/}. Association for Computational
  Linguistics, pages 993--1000.

\bibitem[{Zhang and Zong(2016{\natexlab{a}})}]{zhang2016bridging}
Jiajun Zhang and Chengqing Zong. 2016{\natexlab{a}}.
\newblock Bridging neural machine translation and bilingual dictionaries.
\newblock {\em arXiv preprint arXiv:1610.07272\/} .

\bibitem[{Zhang and Zong(2016{\natexlab{b}})}]{zhang2016exploiting}
Jiajun Zhang and Chengqing Zong. 2016{\natexlab{b}}.
\newblock Exploiting source-side monolingual data in neural machine
  translation.
\newblock In {\em Proceedings of EMNLP\/}.

\end{thebibliography}
\bibliographystyle{ijcnlp2017}

\begin{table*}[t]
\centering
	\begin{subtable}[t]{\columnwidth}
		\centering
		\begin{tabular}{c||cc||cc}
        		\hline
                \textbf{Baseline} & \multicolumn{2}{c||}{Cs \(\rightarrow\) De} & \multicolumn{2}{c}{De \(\rightarrow\) Cs} \\
				(a) Europarl+NC11 & \multicolumn{2}{c||}{13.15} & \multicolumn{2}{c}{11.16} \\ 
                (b) +Pivot back-trans corpus & \multicolumn{2}{c||}{{\scriptsize(+3.82)} 16.97} & \multicolumn{2}{c}{{\scriptsize(+4.24)} 15.40} \\ \hline\hline
				\textbf{Synthetic Corpus} & Pseudo Only & Real Fine-tuning & Pseudo Only & Real Fine-tuning \\
                Cs-De* & 14.77 & {\scriptsize(+1.66)} 16.43 & 14.34 & {\scriptsize(+0.86)} 15.20 \\
                Cs*-De & 16.88 & {\scriptsize(+0.17)} 17.05 & 15.48 & {\scriptsize(+0.53)} \textbf{16.01} \\
                PSEUDO\textsubscript{mix} & 16.98 & {\scriptsize(+0.46)} \textbf{17.44} & 15.66 & {\scriptsize(+0.17)} 15.83 \\
            \hline
		\end{tabular}
		\label{tab:real_finetuning_csde_newstest2011}
        \caption{Cs \(\leftrightarrow\) De}
	\end{subtable}%
	\qquad
    \begin{subtable}[t]{\columnwidth}
		\centering
		\begin{tabular}{c||cc||cc}
        		\hline
                \textbf{Baseline} & \multicolumn{2}{c||}{Fr \(\rightarrow\) De} & \multicolumn{2}{c}{De \(\rightarrow\) Fr} \\
				(a) Europarl+NC11 & \multicolumn{2}{c||}{16.14} & \multicolumn{2}{c}{20.86} \\ 
                (b) +Pivot back-trans corpus & \multicolumn{2}{c||}{{\scriptsize(+1.26)} 17.40} & \multicolumn{2}{c}{{\scriptsize(+1.76)} 22.62} \\ \hline\hline
				\textbf{Synthetic Corpus} & Pseudo Only & Real Fine-tuning & Pseudo Only & Real Fine-tuning \\
                Fr-De* & 15.48 & {\scriptsize(+1.68)} 17.16 & 20.73 & {\scriptsize(+2.07)} 22.80 \\
                Fr*-De & 17.15 & {\scriptsize(+0.54)} 17.69 & 17.60 & {\scriptsize(+5.47)} 23.07 \\
                PSEUDO\textsubscript{mix} & 16.94 & {\scriptsize(+0.95)} \textbf{17.89} & 20.11 & {\scriptsize(+3.11)} \textbf{23.22} \\
            \hline
		\end{tabular}
		\label{tab:real_finetuning_frde_newstest2011}
        \caption{Fr \(\leftrightarrow\) De}
	\end{subtable}
    
\caption{Translation results (BLEU) for Pseudo Only and Real Fine-tuning scenarios evaluated on the newstest 2011 set. For the results of the Real Fine-tuning, the values in parentheses are improvements in BLEU compared to the Pseudo Only setting. The highest BLEU for each translation task is bold-faced.}
\label{tab:real_finetuning_newstest2011}
\end{table*}

\begin{table*}[t]
\centering
	\begin{subtable}[t]{\columnwidth}
		\centering
		\begin{tabular}{c||cc||cc}
        		\hline
                \textbf{Baseline} & \multicolumn{2}{c||}{Cs \(\rightarrow\) De} & \multicolumn{2}{c}{De \(\rightarrow\) Cs} \\
				(a) Europarl+NC11 & \multicolumn{2}{c||}{13.49} & \multicolumn{2}{c}{10.76} \\ 
                (b) +Pivot back-trans corpus & \multicolumn{2}{c||}{{\scriptsize(+3.92)} 17.41} & \multicolumn{2}{c}{{\scriptsize(+4.54)} 15.30} \\ \hline\hline
				\textbf{Synthetic Corpus} & Pseudo Only & Real Fine-tuning & Pseudo Only & Real Fine-tuning \\
                Cs-De* & 15.26 & {\scriptsize(+1.81)} 17.07 & 14.08 & {\scriptsize(+0.79)} 14.87 \\
                Cs*-De & 17.05 & {\scriptsize(+0.13)} 17.18 & 15.17 & {\scriptsize(+0.35)} 15.52 \\
                PSEUDO\textsubscript{mix} & 16.97 & {\scriptsize(+0.57)} \textbf{17.54} & 15.37 & {\scriptsize(+0.28)} \textbf{15.65} \\
            \hline
		\end{tabular}
		\label{tab:real_finetuning_csde_newstest2012}
        \caption{Cs \(\leftrightarrow\) De}
	\end{subtable}%
	\qquad
    \begin{subtable}[t]{\columnwidth}
		\centering
		\begin{tabular}{c||cc||cc}
        		\hline
                \textbf{Baseline} & \multicolumn{2}{c||}{Fr \(\rightarrow\) De} & \multicolumn{2}{c}{De \(\rightarrow\) Fr} \\
				(a) Europarl+NC11 & \multicolumn{2}{c||}{16.36} & \multicolumn{2}{c}{21.45} \\ 
                (b) +Pivot back-trans corpus & \multicolumn{2}{c||}{{\scriptsize(+1.74)} 18.10} & \multicolumn{2}{c}{{\scriptsize(+1.86)} 23.31} \\ \hline\hline
				\textbf{Synthetic Corpus} & Pseudo Only & Real Fine-tuning & Pseudo Only & Real Fine-tuning \\
                Fr-De* & 16.59 & {\scriptsize(+1.23)} 17.82 & 21.56 & {\scriptsize(+1.43)} 22.99 \\
                Fr*-De & 17.42 & {\scriptsize(+0.57)} 17.99 & 18.27 & {\scriptsize(+5.11)} 23.38 \\
                PSEUDO\textsubscript{mix} & 17.42 & {\scriptsize(+0.92)} \textbf{18.34} & 21.20 & {\scriptsize(+2.45)} \textbf{23.65} \\
            \hline
		\end{tabular}
		\label{tab:real_finetuning_frde_newstest2012}
        \caption{Fr \(\leftrightarrow\) De}
	\end{subtable}
    
\caption{Translation results (BLEU) for Pseudo Only and Real Fine-tuning scenarios evaluated on the newstest 2012 set. For the results of the Real Fine-tuning, the values in parentheses are improvements in BLEU compared to the Pseudo Only setting. The highest BLEU for each translation task is bold-faced.}
\label{tab:real_finetuning_newstest2012}
\end{table*}

\end{document}